%% file: main.tex
\documentclass[10pt,twocolumn,letterpaper]{article}

\usepackage[pagenumbers]{iccv} 

\definecolor{iccvblue}{rgb}{0.21,0.49,0.74}
\usepackage[pagebackref,breaklinks,colorlinks,allcolors=iccvblue]{hyperref}

\def\paperID{8961} 
\def\confName{ICCV}
\def\confYear{2025}

\usepackage{graphicx}
\usepackage{amsmath}
\usepackage{physics}
\usepackage{amssymb}
\usepackage{booktabs}
\usepackage{multirow}
\usepackage[accsupp]{axessibility}
\usepackage{calc}
\usepackage{mathtools}
\usepackage[capitalize]{cleveref}
\usepackage[nolist]{acronym}
\usepackage[numbers]{natbib}
\usepackage{collect}
\usepackage{flushend}
\usepackage{adjustbox}
\usepackage{bm}
\usepackage{algorithm}
\usepackage{algpseudocode}
\usepackage{siunitx}
\usepackage[accsupp]{axessibility}
\usepackage{xspace}
\usepackage{collcell}
\usepackage{mhchem}
\usepackage{tabularx}
\usepackage{svg}
\usepackage{array}
\usepackage{xfrac}

\usepackage[accsupp]{axessibility}  

\DeclareMathOperator*{\E}{\mathbb{E}}

\input{tr-commands}

\begin{document}

\input{project-commands}

\makeatletter
\newcommand{\subalign}[1]{%
  \vcenter{%
    \Let@ \restore@math@cr \default@tag
    \baselineskip\fontdimen10 \scriptfont\tw@
    \advance\baselineskip\fontdimen12 \scriptfont\tw@
    \lineskip\thr@@\fontdimen8 \scriptfont\thr@@
    \lineskiplimit\lineskip
    \ialign{\hfil$\m@th\scriptstyle##$&$\m@th\scriptstyle{}##$\hfil\crcr
      #1\crcr
    }%
  }%
}
\makeatother

\title{Generative Video Bi-flow}

\author{Chen Liu\\
University College London\\
{\tt\small chen.liu.21@ucl.ac.uk}
\and
Tobias Ritschel\\
University College London\\
{\tt\small t.ritschel@ucl.ac.uk}
}
\maketitle

\begin{abstract}
We propose a novel generative video model to robustly learn temporal change as a neural \ac{ODE} flow with a bilinear objective which combines two aspects:
The first is to map from the past into future video frames directly. 
Previous work has mapped the noise to new frames, a more computationally expensive process.
Unfortunately, starting from the previous frame, instead of noise, is more prone to drifting errors.
Hence, second, we additionally learn how to remove the accumulated errors as the joint objective by adding noise during training.
We demonstrate unconditional video generation in a streaming manner for various video datasets, all at competitive quality compared to a conditional diffusion baseline but with higher speed, \ie fewer \ac{ODE} solver steps.
\end{abstract}

\vspace{-.1cm}
\mysection{Introduction}{introduction}
We suggest a method to generate videos.
Our model is based on a neural \ac{ODE} flow, modeling the pixel evolution over time.
To train the model, we reformulate the flow matching objective \cite{lipmanFlowMatchingGenerative2022, liuFlowStraightFast2022, heitzIterativeDeBlendingMinimalist2023, albergoBuildingNormalizingFlows2023}, which has been widely used to map the noise distribution to an image or video distribution.
One of our key insights is to consider a video dataset as separate image datasets in the temporal dimension.
We can now learn flow between the current and the next-frame distribution.
To make this work, we need to improve the coverage of the training space and learn a joint flow field of denoising by using train-time noise.
This enables robust inference with few \ac{ODE} solver steps.
Our approach can produce new video frames in a streaming fashion at an interactive rate, making it a candidate technique for interactive applications such as computer games.

\begin{figure}[htb]
\centering\includegraphics*[width = \linewidth]{\figurePath Alternatives}
\vspace{-.6cm}
\caption{The two main ways to generate videos on the left and our approaches on the right.
The rightmost is our final approach.
``Coverage'' denotes if training sees all conditions occurring at test.
Higher coverage indicates more robust inference, while low coverage results in quick divergence.
``Distance'' is the length of the \ac{ODE} solver path. 
The shorter distance to solve means fewer steps.
``Streaming'' is the ability to adapt to the condition and produce an infinite image stream using finite memory.
}
\vspace{-.6cm}
\label{fig:Alternatives}
\end{figure}

So far, there are mainly two dominant ways to generate videos, which we compare to our approaches, summarized as different columns in \refFig{Alternatives}.
The first approach, full-sequence diffusion \cite{hoVideoDiffusionModels2022}, is a straightforward extension from images to videos.
With large curated video datasets and established methods such as the noise conditional score function \cite{songGenerativeModelingEstimating2019}, this achieves good coverage of empirical video distributions, denoted as a success in the row ``High coverage'' in \refFig{Alternatives}.
However, it is a computationally intense process in both memory and time as the diffusion problem requires numerous steps to solve from a 3D noise tensor to a complete video, denoted as a challenge in the ``Short distance'' row in \refFig{Alternatives}.
It also does not support streaming video generation or the ability to react to external input in interactive applications.
We seek to improve these aspects.

The second approach, conditional diffusion, is performed on frames in isolation.
To help temporal consistency, the generation is conditioned on one or multiple previous frames.
This allows for streaming, but still starts from noise and requires solving a large distance, as a result of which the compute demands remain high.

We introduce here a third approach, \emph{video streaming flow}, which relates to the classic single-image animation problem, generating a plausible video sequence given a single image.
The \ac{ODE} flow models the gradient concerning the current frame and predicts the next frame directly from it, thus eliminating the lengthy denoising process.
Such an approach is ideal, as it allows streaming, and does not need to change much between frames.
Nevertheless, since the flow is trained solely to map from one frame to another rather than mapping noisy inputs to clean frames, its coverage of the training space of temporal variations is limited.
Consequently, the process becomes unstable, and the solution diverges after only a couple of frames.

Our final approach, \emph{video bi-flow}, combines the strengths of both the second and the third approach.
It infers each frame directly from the previous one, without starting from noise, but still injects sufficient noise during training to adequately cover the space of transitions and ensure stable solving.
The resulting method is both computationally efficient and capable of stable streaming.

In summary, our contributions are:
\begin{itemize}
    \item Learning of video flow to exploit correlation between frames to enable efficient streaming video generation;
    \item Mitigating error accumulation by a novel joint flow field, facilitating stable and long video generation;
    \item Experiments to systematically demonstrate that our video bi-flow exhibits superiority over the conditional diffusion both in stability and efficiency with comparable fidelity.
\end{itemize}
Our code will be released at \url{https://github.com/ryushinn/ode-video}.

\mysection{Our Approach}{OurApproach}

\mysubsection{Joint objective}{Objective}

Flow matching \cite{lipmanFlowMatchingGenerative2022, liuFlowStraightFast2022} aims to regress a flow field $\ode$ to a pre-defined probability path between two distributions $\distOne$ and $\distTwo$:
\begin{align}
    \label{eq:iadb}
    \argmin{\parameters} \E_{\sourceSample, \targetSample, \noiseCoord} \left||\ode(\sample_\noiseCoord,\noiseCoord) - (\targetSample - \sourceSample)\right||^2,
\end{align}
where $\sourceSample \sim \distOne, \targetSample \sim \distTwo$, $\noiseCoord \sim \uniformDistribution(0, 1)$, and $\sample_\noiseCoord=\sourceSample + \noiseCoord(\targetSample - \sourceSample)$ is linear interpolation, such that it can drive a sample $\sourceInfSample $ from $\distOne$ to a sample $\targetInfSample$ following $\distTwo$:
\begin{align}
\label{eq:odeSolution}
\targetInfSample
=
\solve_{0\to 1}(\sourceInfSample, \ode)
=
\sourceInfSample +
\int_{0}^{1}
\ode(\infSample_\solveCoord, \solveCoord) \dd \solveCoord,
\end{align}
where $\solve$ is an off-the-shelf \ac{ODE} solver that solves $\ode$ with the initial value $\sourceInfSample$ from $0$ to $1$.

This method is widely applied in image generation which we will first review. 
We then reformulate it to video generation by mapping from the distributions of past frames to the distribution of future frames. 
We finally introduce the robust training of our video flow, the key to our approach, where we combine the benefits of the above two formulations with bi-linear interpolation.

\mycfigure{Concept}{
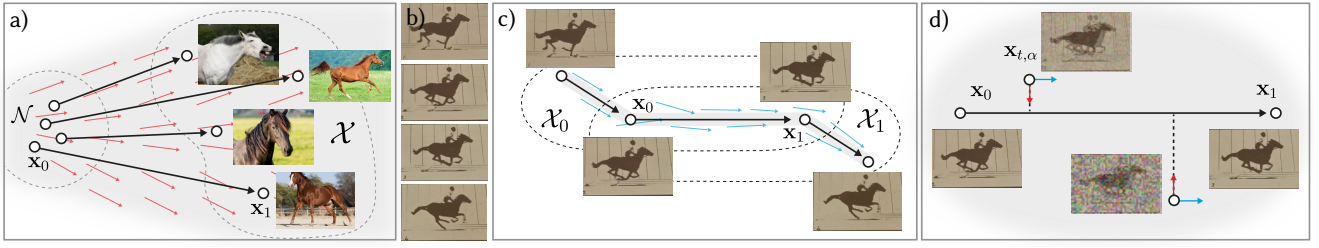
The main idea of our approach:
We extend the flow matching shown in \textbf{a)} which learns a flow from a normal distribution to a distribution of images.
Consider a sequence of four video frames in \textbf{b)}.
\textbf{c)} shows the sequence and the linear interpolations between them as black lines in the respective order.
Note that the horizontal axis is not time, but time flows along the line.
We learn a flow field to drive these trajectories by applying flow matching between two consecutive frame distributions $\distOne$ and $\distTwo$.
The fitted flow field is, unfortunately, only stable in a small region around these linear connections, shown as the gray area.
\textbf{d)} shows our final approach, for simplicity zooming onto a link of only two frames.
We add noise to the interpolated points during training, and two random points are shown as examples.
The flow field is then tasked to map these distorted points forward (blue arrows), but also denoise them back to clean interpolations (red arrows).
This widens the coverage of the training frame space, as shown as the populated gray area, resulting in a flow field that can effectively recover from errors accumulating over many frames.
}

\vspace{-.1cm}
\paragraph{Image generation}
To generate images, $\distOne$ is usually a normal distribution $\normalDistribution$ that we can sample, and $\distTwo$ is an empirical image distribution formed by the image data of interest.
The \ac{ODE} field trained by solving \refEq{iadb} is the expectation of all possible linear paths between random pairs of noises and images, as depicted by \refFig{Concept}a.
In inference, a new image can be generated by solving \refEq{odeSolution} with a normal-random initial condition.

\paragraph{Video generation}
A video distribution can be formalized as temporally evolving distributions of image frames. 
The transport describing this manifold morphing provides a mechanism for generating videos, and the key challenge is deriving such a transport from data.
To this end, we define \distOne as the image distribution of all frames in a video dataset and \distTwo as the image distribution composed by all the next frames of every image from \distOne.
Let $\pair(\distOne,\distTwo)$ be the joint distribution of all these consecutive frame pairs; this gives us an empirical approximation of our target transport.

By flow matching, we can then regress a neural \ac{ODE} flow to this target transport by minimizing:
\begin{align}
\label{eq:videoflow}
\argmin{\parameters} 
\E_{(\sourceSample, \targetSample), \timeCoord}
||\ode(\sample_\timeCoord, \timeCoord) - (\targetSample - \sourceSample)||^2,
\end{align}
where $\timeCoord \sim \uniformDistribution(0, 1)$, $(\sourceSample, \targetSample)$ is sampled from $\pair(\distOne, \distTwo)$, and $\sample_{\timeCoord}=\sourceSample + \timeCoord(\targetSample - \sourceSample)$. \refFig{Concept}c illustrates this flow.

Specifically, we minimize to transition from one image distribution to another.
Instead of using independently random data pairs from $\distOne$ and $\distTwo$, we employ $\pair$ during training, the intrinsic coupling in video data.
A proper dependent coupling can result in straighter and easier-to-solve trajectories as shown by \citet{liuFlowStraightFast2022}.
If solved exactly in \refEq{videoflow}, our coupling provides a valid \ac{ODE} flow field driving $\distOne$ to $\distTwo$.
By iteratively solving the learned \ac{ODE}, we can generate an entire trajectory of coherent frames as a video.

Compared to conditional diffusion models which generate the next frame by starting from noise and conditioning the last frame, we show that our video flow significantly reduces the number of steps to stream the video generation with comparable quality and consistency.
Intuitively, it is because we start from an initial state closer to the solution when rolling out the next frame.

\paragraph{Bi-linearity for robustness}
Unfortunately, the video flow alone seems not to be sufficiently stable to advance in time across many frames.
The original approach for images maps between noise and distribution of images, while here, we need to map between images and do so in a stable fashion, \ie iterating as many times as the video has frames.
In this condition, the key problem is the accumulation of errors, and the inability to recover from these during inference, as they were never observed during training.

As in \refEq{videoflow}, $\sample_\timeCoord$ lies in the image manifold and its linear interpolations, the training of the video flow can only supervise a rather low coverage of the entire space, in contrast to the common flow matching which blends noise and image data during training.
This results in that the video flow is sensitive to any accumulated errors and deviates quickly.

As a remedy, we populate the training space with noise and bi-linear interpolation and train a joint flow field called \emph{video bi-flow} which predicts both the video direction and the denoising direction.

We denote a bi-linear interpolation
\begin{align}
    \sample_{\timeCoord,\noiseCoord} = \sourceSample + \timeCoord(\targetSample-\sourceSample) + \alpha \noiseSample
    ,
\end{align}
which first blends two consecutive frames by time factor \timeCoord, and then blends the result with random noise $\noiseSample \sim \normalDistribution$ of magnitude $\noiseCoord$. 
The noise here serves as a surrogate of all potential artifacts accumulated during inference, given that we have no easy access to their true distribution.
Combining \refEq{iadb} and \refEq{videoflow}, the training objective now becomes:
\begin{equation}
\label{eq:bivideoflow}
\begin{aligned}
    \argmin{\parameters} 
    \E_{\substack{
        (\sourceSample, \targetSample), \noiseSample, \\
        \timeCoord, \noiseCoord
    }}
    \quad[ &||
        \odev(\sample_{\timeCoord,\noiseCoord}, \timeCoord, \noiseCoord)
        - (\targetSample - \sourceSample)
    ||^2
    \\
    + \quad &||
        \oden(\sample_{\timeCoord,\noiseCoord}, \timeCoord, \noiseCoord)
        - \noiseSample
    ||^2].
\end{aligned}
\end{equation}
where the target $\ode=(\odev, \oden)$ is indeed a joint field of two \ac{ODE} flow fields of moving forward in time and denoising, as depicted in \refFig{Concept}d.
The video field $\odev$, as optimized in the first term, predicts the-next-frame trajectory under a noise level of \noiseCoord and the (de)noising field $\oden$, as optimized in the second term, predicts the noising errors added and can drive the corrupted images to clean images or their linear mix by $\timeCoord$.
Thus, the video bi-flow subsumes the two cases of generation mentioned above as two extremes when $\noiseCoord=0$ or $\timeCoord \in \{0, 1\}$.

Our bi-linear training objective achieves the same aim as described by \citet{songGenerativeModelingEstimating2019} for classic diffusion: to increase the coverage of the training space -- for us video frames -- therefore stabilizing the generation process. 

\mysubsection{Joint sampling patterns}
{Sampling}

In inference, our trained video bi-flow can be solved along the dimension of $\noiseCoord$ as the denoiser/corrector:
\begin{align}
    \label{eq:generator}
    \frac{\partial \sample_{\timeCoord, \noiseCoord}}{\partial \noiseCoord} = \oden(\sample_{\timeCoord, \noiseCoord}, \timeCoord, \noiseCoord),
\end{align}
or along the dimension of $\timeCoord$ as the predictor:
\begin{align}
    \label{eq:predictor}
    \frac{\partial \sample_{\timeCoord, \noiseCoord}}{\partial \timeCoord} = \odev(\sample_{\timeCoord, \noiseCoord}, \timeCoord, \noiseCoord).
\end{align}

In practice, we evolve one given frame with the predictor to form a video. 
A frame generated by a separate model or a random frame in the test set can be the first frame.
To ensure stable video streaming, we additionally employ the denoiser to remove the error accumulated at each step with a joint sampling pattern, which is the characteristic curve \cite{pinchoverIntroductionPartialDifferential2005} to solve the first-order \ac{PDE} field $\ode$ in the space of $(\timeCoord, \noiseCoord)$, along which the \ac{PDE} becomes an \ac{ODE}.
Below, we will review the streaming generation and introduce our joint sampling pattern.

\paragraph{Streaming generation}

We aim to sample a video of $\numberOfFrames$ frames \mbox{$\{\infSample^{\frameSub} | \frameSub=0,1,...,\numberOfFrames - 1\}$} with the first frame $\infSample^{0}$ given.
The following frames are generated by Markovian sampling, where we solve \refEq{predictor} to get $\infSample^{\frameSub}=\solve_{0\to 1}(\infSample^{\frameSub-1}, \odev(\cdot, \cdot, 0))$.
Note that here we set $\noiseCoord=0$, \ie we have to assume the last frame is clean without reliable prior information on how many errors have been accumulated.
In this pattern, the denoising flow field $\oden$ in \refEq{generator} is never used, but we will show that it is an effective corrector to significantly mitigate the accumulated drift.

\paragraph{Joint sampling}
The joint modeling actually enables one to solve simultaneously in both the temporal and denoising dimensions.
Our bilinear video flow is a first-order \ac{PDE} system in the $\timeCoord\noiseCoord$-plane, where we can consider a characteristic curve $(\timeCoord_\jointCoord, \noiseCoord_\jointCoord)$ parameterized by $\jointCoord \in [0, 1]$. 
Solving the \ac{PDE} at $(\timeCoord_1, \noiseCoord_1)$ with an initial value at $(\timeCoord_0, \noiseCoord_0)$ reduces to solving an \ac{ODE} along this curve.
The corresponding characteristic \ac{ODE} of this curve is:
\begin{equation}
    \label{eq:joint}
    \begin{aligned}
    &\frac{\dd \sample_{\timeCoord_\jointCoord,\noiseCoord_\jointCoord}}{\dd \jointCoord} = \frac{\partial \sample}{\partial \timeCoord}\frac{\dd\timeCoord}{\dd\jointCoord} +  \frac{\partial \sample}{\partial \noiseCoord}\frac{\dd\noiseCoord}{\dd\jointCoord} =\\ 
    &
    \underbrace{
        \odev(
        \sample_{\timeCoord_\jointCoord,\noiseCoord_\jointCoord}, \timeCoord_\jointCoord, \noiseCoord_\jointCoord
        )
        \frac{\dd \timeCoord_\jointCoord}{\dd \jointCoord}
        +
        \oden(
        \sample_{\timeCoord_\jointCoord,\noiseCoord_\jointCoord}, \timeCoord_\jointCoord, \noiseCoord_\jointCoord
        )
        \frac{\dd \noiseCoord_\jointCoord}{\dd \jointCoord}
        }_{\odej(
        \sample_{\timeCoord_\jointCoord,\noiseCoord_\jointCoord}, \jointCoord
        )
    }
    .
    \end{aligned}
\end{equation}

Thereafter, we can achieve the correction and evolution jointly by solving \refEq{joint}:
\begin{equation}
    \label{eq:joint_solve}
\infSample^{\frameSub} = \solve_{0\to 1}(\infSample^{\frameSub-1} + \noiseLevel\noiseSample, \odej),
\end{equation}
where we set as the initial value the last generated frame with noise added, and solve along a curve from $(\timeCoord_0, \noiseCoord_0)=(0, \noiseLevel)$ to $(\timeCoord_1,\noiseCoord_1)=(1, 0)$, \ie advancing to the next frame while applying denoising.
The added noise simulates the drifting error over time such that we can explicitly remove it, yielding a more stable video trajectory compared to the na\"ive streaming generation.
The noise level $\noiseLevel$ is a hyperparameter that we can tune to trade off video consistency for frame quality during inference, and we study its different levels in the evaluation.

We only consider the linear curve straightly connecting two endpoints so $(\frac{\dd \timeCoord_\jointCoord}{\dd \jointCoord}, \frac{\dd \noiseCoord_\jointCoord}{\dd \jointCoord})=(1, -\noiseLevel)$.
We find experimentally that it works better than two alternative nonlinear curves, the solve-then-denoise curve ($(\timeCoord_{0.5}, \noiseCoord_{0.5})=(1, \noiseLevel)$) and the denoise-then-solve curve ($(\timeCoord_{0.5}, \noiseCoord_{0.5})=(0, 0)$).

\mysection{Experiments}{Experiments}
\mysubsection{Protocol}{Protocol}
\paragraph{Dataset}
We extensively validate our video bi-flow in six video datasets with varying levels of variations: 
\dataset{MINERL} (Minecraft gameply videos of an agent \cite{saxenaClockworkVariationalAutoencoders2021}), 
\dataset{MAZES} (videos of navigating a maze \cite{saxenaClockworkVariationalAutoencoders2021}),
\dataset{CARLA} (car driving videos \cite{harveyFlexibleDiffusionModeling2022}),
\dataset{SKY} (time-lapse videos of sky \cite{xiongLearningGenerateTimeLapse2018}),
\dataset{BIKING} (first-person view mountain biking footage \cite{brooksGeneratingLongVideos2022}),
and \dataset{RIDING} (horse riding videos \cite{brooksGeneratingLongVideos2022}).
We resize the videos to $128^2$ for training, except for \dataset{MINERL} and \dataset{MAZES}, for which we use their original resolution of $64^2$.
We use the default train-test split for all datasets: all methods are trained in the train set and then used to generate videos given random frames from the test set as the starting points.
We adopt a random 80-20 split for those datasets without a default split.
We only experiment with unconditional video generation.

\paragraph{Metrics}
We quantify our results in terms of quality and compute speed.
The quality metric is \ac{FVD} \cite{Unterthiner2019FVDAN}.
We measure \ac{FVD} on 128 samples of 512 frames generated by each method against the test set.
To evaluate the quality change over time, the \ac{FVD} is measured with a sliding window of 32 frames in a stride of 16 frames.
The compute speed is simply the wall-clock time used to generate one frame. 
It is measured in the unit of the number of necessary steps\footnote{Precisely, it is the number of \ac{ODE} network evaluations.} to solve the trained flow by an adaptive \ac{ODE} solver under the same accuracy.

We show more numerical analysis with frame-based metrics in the supplemental.

\paragraph{Methods}

Our baseline method is a conditional diffusion model generating a frame conditioned on the previous frame, as referred to as \method{ConDiff}\footnote{With a slight abuse of terminology, it is in fact a conditional flow model, \ie a deterministic diffusion model.}, which we compare to our video bi-flow (\method{Bi-flow}) and its ablation: training our proposed video streaming flow without the bilinear extension (\method{Flow}).

We train all methods with the same network architecture using the same setting and hyper-parameters, only differing in the losses and the throughput.
Specifically, \method{ConDiff} is trained by the conditional version of \refEq{iadb} with the last frame concatenated to noise as the condition. 
Thus, \method{ConDiff} yields a throughput of $6\rightarrow3$, \ie taking a six-channel input and predicting the three-channel denoising gradient.
\method{Flow} is trained by \refEq{videoflow} with the throughput of $3\rightarrow3$.
\method{Bi-flow} is trained by \refEq{bivideoflow}, our bi-linearity loss, to enable fast and stable video streaming. 
As a joint field, \method{Bi-flow} has a throughput of $3\rightarrow 6$: each half of the six-channel output is the partial derivative along time or (de)noising.
We use a 2D UNet without any dedicated temporal modeling.

In inference, we solve the learned flows using an adaptive Heun \ac{ODE} solver with an absolute and relative tolerance of $10^{-2}$.
For \method{ConDiff}, we iteratively use the last frame as the condition and solve the trajectory from random noise to the next frame.
In contrast, \method{Flow} and \method{Bi-flow} solve the trajectory from the last frame to the next directly, while \method{Bi-flow} further employs joint sampling (\refEq{joint_solve})
 
We conducted all experiments on the Nvidia RTX 4090 GPU.
More implementation details can be found in the supplemental material.

\begin{table*}[]
    \centering
    \caption{The \textbf{table} reports the main quantitative results of our and other approaches (rows) for different video datasets (columns). Numbers here are averaged over time;
    The \textbf{scatter plots} below show how quality and speed relate: an ideal method would have a low step count and a low \ac{FVD} to reside in the bottom-left corner.
    We also plot \method{Bi-flow} with varying levels of the noise \noiseLevel applied in sampling, as 0.0, 0.1, 0.2, and 0.3.
    These variants form the Pareto front of our \method{Bi-flow} connected by dots, depicting the spectrum of performance we can alternatively favor based on the application.
    Note that it is our proposed joint flow field that enables this trade-off dimension, independent of the \ac{ODE} solver.
    The noise level yielding the best \ac{FVD} for each dataset is marked with \emph{a star} in the scatter plot.
    This variant is reported in the table and also henceforth reported by default in other evaluations;
    The \textbf{line charts} are the linear trend lines of \ac{FVD} vs. Time, plotted by computing the \acp{FVD} in a sliding window on the entire video.
    Here ``Time'' denotes the frame index.
    A flat trend line indicates a method can effectively mitigate the accumulated errors and maintain video quality over time.
    A positive slope means error accumulation.
    }
    \label{tab:main}
    \resizebox{\textwidth}{!}{%
    \begin{tabular}{crrrrrrrrrrrrr}\toprule
    &\multicolumn{2}{c}{\dataset{SKY}} &\multicolumn{2}{c}{\dataset{BIKING}} &\multicolumn{2}{c}{\dataset{RIDING}} &\multicolumn{2}{c}{\dataset{CARLA}} &\multicolumn{2}{c}{\dataset{MAZES}} &\multicolumn{2}{c}{\dataset{MINERL}} \\
    \cmidrule(lr){2-3}\cmidrule(lr){4-5}\cmidrule(lr){6-7}\cmidrule(lr){8-9}\cmidrule(lr){10-11}\cmidrule(lr){12-13}
    &\lessIsBetter{FVD} &\lessIsBetter{Steps} &\lessIsBetter{FVD} &\lessIsBetter{Steps} &\lessIsBetter{FVD} &\lessIsBetter{Steps} &\lessIsBetter{FVD} &\lessIsBetter{Steps} &\lessIsBetter{FVD} &\lessIsBetter{Steps} &\lessIsBetter{FVD} &\lessIsBetter{Steps} \\\midrule
    \method{Flow} &2144 &\winner{7.84} &14010 &\winner{10.38} &9154 &\winner{7.43} &4913 &\winner{7.78} &5051 &\winner{10.02} &3259 &\winner{10.20} \\
    \method{Bi-flow} &\winner{526} &14.57 &\winner{3753} &25.63 &\winner{1552} &13.73 &\winner{345} &23.51 &605 &29.62 &\winner{834} &25.73 \\
    \method{ConDiff} &1271 &46.18 &11465 &33.15 &4668 &37.73 &947 &53.77 &\winner{424} &45.31 &1149 &41.61 \\
    \bottomrule
    \end{tabular}
    }
    	
    \includegraphics[width=\textwidth]{\figurePath/Table}
\end{table*}

\newcommand{\comphead}[1]{\multicolumn1c{\dataset{\scriptsize{#1}}}}

\begin{table}[!htp]\centering
\caption{\ac{FVD} reported for comparisons between methods under the same number of steps.}\label{tab:equal_time}
\renewcommand{\tabcolsep}{0.1cm}
\begin{tabular}{lrrrrrrr}
\toprule
&
\comphead{SKY}&
\comphead{BIKING}&
\comphead{RIDING}&
\comphead{CARLA}&
\comphead{MAZES}&
\comphead{MINERL}
\\\midrule
\method{Flow} &2109 &11158 &8984 &4931 &4928 &3170 \\
\method{Bi-flow} &\winner{491} &\winner{3271} &\winner{1621} &\winner{307} &\winner{620} &\winner{823} \\
\method{ConDiff} &994 &10253 &6698 &829 &655 &2286 \\
\bottomrule
\end{tabular}
\end{table}

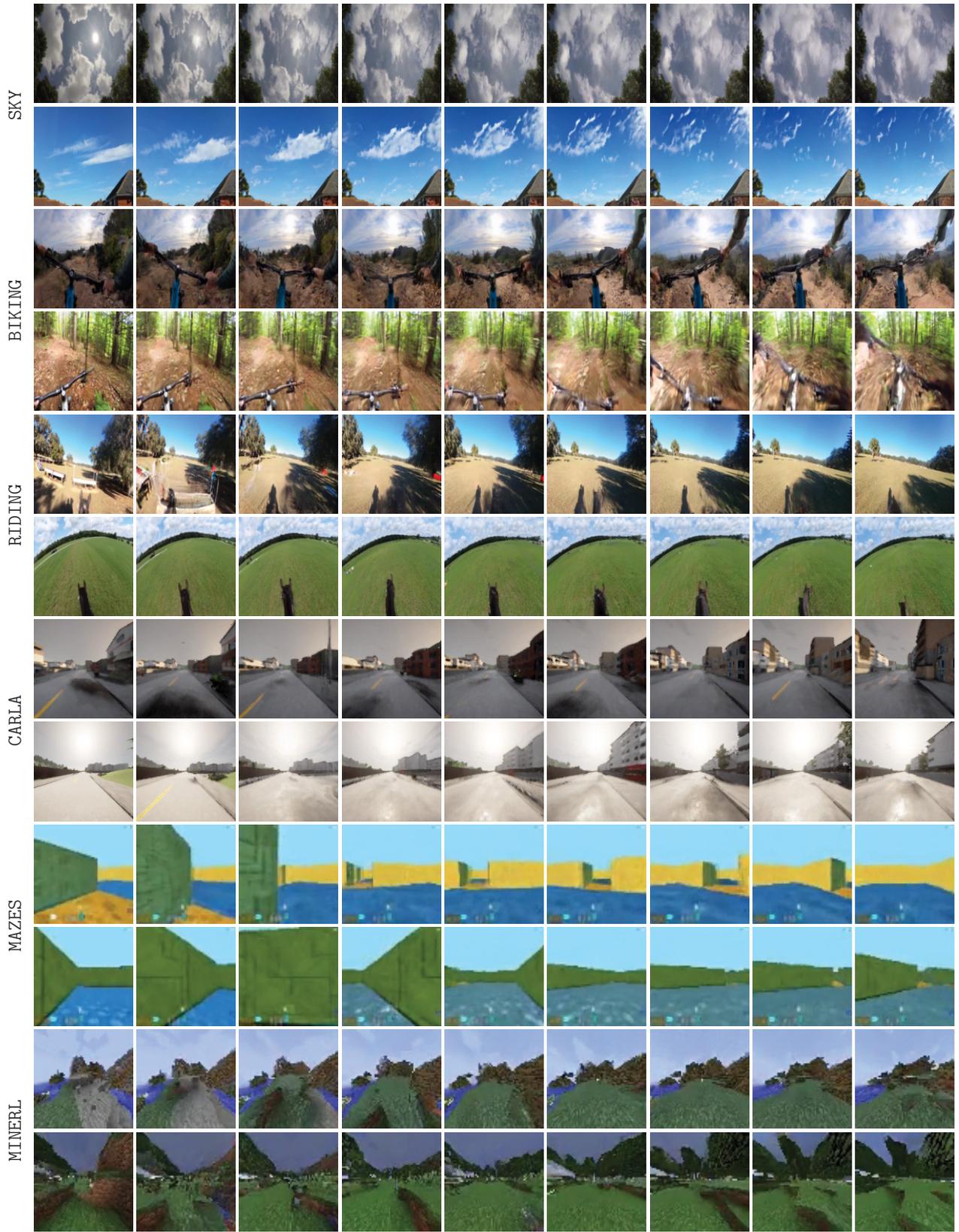
\mycfigure{Qualitative}{Videos samples generated by \method{Bi-flow}. Time goes from left to right.}
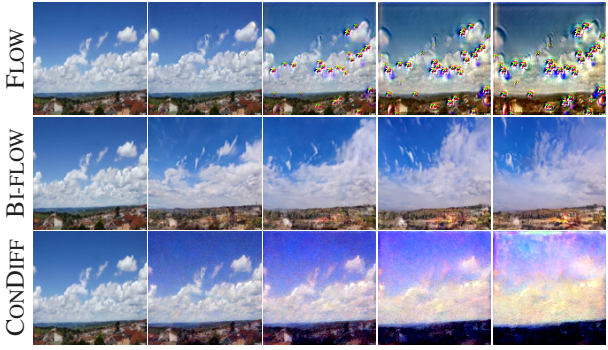
\myfigure{QualitativeComparison}{Five frames generated by different methods for \dataset{SKY}. These five frames uniformly span 128 frames in total, four times longer than 32-frame training data. Our \method{Bi-flow} maintains the best quality throughout, while other methods induce obvious artifacts near the end of the video.}

\mysubsection{Evaluation}{Evaluations}
\paragraph{Quantitative}
The main quantitative results are seen in \refTab{main}.
\method{Flow} requires the lowest number of steps per frame, but its solving trajectory rapidly diverges due to a lack of error mitigation, resulting in the highest \ac{FVD}.
In contrast, \method{ConDiff} can achieve decent \acp{FVD}, albeit at the expense of a substantially increased step count, as it always starts from noise when it rolls out the next frame.
As illustrated in the scatter plots, these two methods lie in the top-left and bottom-right corners, respectively.
\method{Bi-flow}, however, generates videos robustly with denoising flow as the corrector.
Compared to \method{ConDiff}, \method{Bi-flow} consistently attains superior---at least comparable---\acp{FVD} across all datasets while using fewer necessary steps through the fast video flow predictor.
Notably, \method{Bi-flow} occupies the ideal bottom-left quadrant in all scatter plots, indicating its ability to synthesize high-quality videos with minimal computational overhead.

In plots of \refTab{main}, we also evaluate four different noise levels \noiseLevel during the joint sampling, ranging from 0.0 to 0.3, annotated with four gradations of blue to reflect the increase in noise levels.
We find that increasing \noiseLevel requires slightly more solving steps while taming error accumulation, as evidenced by reduced slopes in the line charts:
\method{Bi-flow} with $\noiseLevel=0.0$ (\ie using the na\"ive streaming generation) exhibits a sharp \ac{FVD} rise akin to \method{Flow} and \method{ConDiff}, but presents nearly flat trend lines with higher \noiseLevel.
Thus, it is necessary to adopt an appropriate \noiseLevel for stable long video generation.

However, increasing $\noiseLevel$ does not necessarily decrease the average \ac{FVD} over time, despite enhancing long-term stability, as it is also a trade-off between the intra-frame quality and the inter-frame consistency.
Although higher \noiseLevel enables stronger denoising corrections that improve per-frame fidelity and suppress error propagation, it paradoxically degrades temporal coherence due to the additional noise introduced.
Therefore, depending on the emphasis on steps, stability, or the video quality, the optimal \noiseLevel can vary vastly across datasets.
Here, we report the noise level yielding the best \ac{FVD} by default for each dataset since it usually achieves a good balance among these three aspects with only half of steps compared to \method{ConDiff} on average.

Besides the results of equal-accuracy by an adaptive \ac{ODE} solver, we also conduct equal-time experiments by an Euler \ac{ODE} solver with a fixed step size of $0.05$, that is all methods operate under the same budget of 20 steps, as shown in \refTab{equal_time}. 
\method{Bi-flow} consistently outperforms \method{ConDiff} in this scenario.

\paragraph{Qualitative}
We show frames generated by \method{Bi-flow} in \refFig{Qualitative}.
Using the denoising flow, our method can produce decent-quality long videos without noticeable degradation over time.
Notably, for \dataset{SKY} dataset comprising videos with only 32 frames, our method can even robustly stream videos significantly longer than the training data, \ie achieving video extrapolation.
We also qualitatively compare our methods and baselines in \refFig{QualitativeComparison}, which demonstrates the effectiveness of our joint flow design in removing accumulated errors.

We present more videos for all methods and datasets in the supplemental.

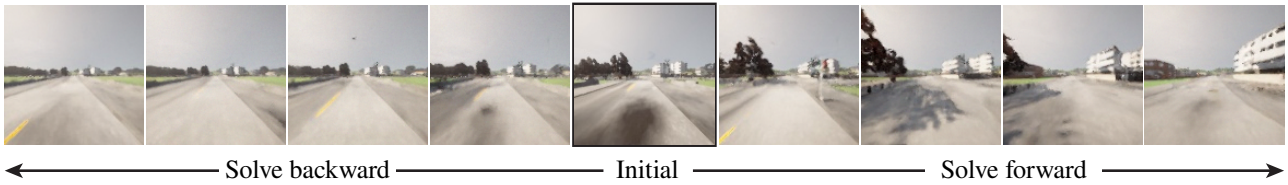
\mycfigure{SolveBackward}{The given initial frame is in the middle. Frames on either side are solved with our video bi-flow forward or backward, respectively.}

\mysection{Related Works}{RelatedWork}
\mysubsection{Video diffusion models}{VideoDiffusionModels}
Diffusion models \cite{hoDenoisingDiffusionProbabilistic2020} initially applied its success to the video domain by directly expanding states to 3D tensors and adopting (factorized) 3D convolutions and attentions \cite{hoVideoDiffusionModels2022, blattmannAlignYourLatents2023}.
To resolve the high compute demand associated with large 3D tensors, hierarchical synthesis and latent modeling are usually essential \cite{blattmannStableVideoDiffusion2023, bar-talLumiereSpaceTimeDiffusion2024, hongCogVideoLargescalePretraining2022, jin2025pyramidal, girdharEmuVideoFactorizing2023}, extending not only across spatial dimensions \cite{rombachHighResolutionImageSynthesis2022} but also temporally \cite{hoImagenVideoHigh2022}, especially when long-duration video generation is a primary objective \cite{harveyFlexibleDiffusionModeling2022, brooksGeneratingLongVideos2022, heLatentVideoDiffusion2023}.
Nevertheless, hierarchical video synthesis is not suitable for interactive applications since users have to wait until an entire full sequence of video has been generated.
Alternatively, generating videos auto-regressively in the temporal dimension has been investigated in prior works \cite{hoVideoDiffusionModels2022, harveyFlexibleDiffusionModeling2022} as another conditioning option.
One of the most recent works is GameNGen \cite{valevski2025diffusion} which runs a diffusion model auto-regressively to generate videos, conditioned on the user inputs and all the past generated frames up to the memory budget, and simulates an infinite environment for the video game ``Doom''.

We refer to this paradigm as conditional diffusion, irrespective of the conditioning mechanism employed.
They all diffuse from noise.
Our method, in contrast, addresses the inherent lengthy sampling process of diffusion models by commencing our video flow directly from the previous frame rather than random noise, thereby yielding faster sampling.

LVDM \cite{blattmannAlignYourLatents2023} and GameNGen \cite{valevski2025diffusion} also adopt noise to perturb conditions during training for robust inference.
Diffusion forcing \cite{chenDiffusionForcingNexttoken2024} presents a similar idea that their diffusion model can condition on any noisy version of previous frames, ranging from prior (pure noise as a condition) to posterior (a clean image as a condition). 
They achieve this by training a separate recurrent network to update the state token based on noisy observations, on which the image diffusion model will condition. 
Our method not only augments the input with noise but also learns to remove the added noise explicitly as one branch of the joint flow field.
The learned joint flow field builds a first-order \ac{PDE} in the 2D space of time and denoising, which supports flexible sampling paths, such as the efficient and robust joint sampling pattern (\refSec{Sampling}) and backward sampling (\refSec{Discussion}).

Recent works share our observation that the last frame is a better starting point for solving the next frame.
RIVER \cite{davtyan2023efficient} accelerates the sampling process by truncated denoising, which solves the denoising ODE with the noisy last frame as the initial value and a reduced integration range. 
SVP \cite{ostrek2024stable}, primarily for smoother results rather than acceleration, performs a weighted sum of the current state, the last frame, and noise at an intermediate denoising step.
They can be considered as special cases of conditional diffusion with modified inference procedures.
Our fundamental difference from them is the \ac{ODE} of time, which we show is a much more effective flow to generate the next frame with fewer steps and deserves dedicated modeling.

\mysubsection{Diffusion bridges and coupled data}{Bridge}
A recent family of methods named diffusion bridges focuses on generalizing the diffusion models to any two arbitrary distributions, rather than relying on the non-informative Gaussian as the prior distribution \cite{zhouDenoisingDiffusionBridge2023, bortoliAugmentedBridgeMatching2023, liuI$^2$SBImagetoImageSchrodinger2023, shiDiffusionSchrodingerBridge2023}.
By leveraging the correlation between distributions, especially the coupled data pairs, \eg in the image-to-image translation task, the bridge models exhibit superior performance compared to the traditional diffusions.
The same idea has also been adopted and explored in flow models.
\citet{heitzIterativeDeBlendingMinimalist2023} investigate some data-to-data applications such as colorization and super-resolution.
\citet{liuFlowStraightFast2022} show that we can iteratively straighten the transport to achieve faster sampling by a new round of training based on the last learned coupling, or, in the first iteration, the independent coupling.

Inspired by the potential of coupled data in these successes, we propose leveraging the inherent inter-frame correlation in video data, an aspect that has yet to be sufficiently explored, and demonstrate that it represents a more efficient and favorable transport flow toward the subsequent frame.

\mysubsection{Neural differential equations for videos}{node}
\citet{chenNeuralOrdinaryDifferential2018} propose neural \ac{ODE} with the adjoint gradient method.
Flow matching \cite{lipmanFlowMatchingGenerative2022} is a simulation-free training method for neural \ac{ODE} to learn an \ac{ODE} flow that transforms one distribution to another distribution.
As revealed by recent works \cite{heitzIterativeDeBlendingMinimalist2023, liuFlowStraightFast2022} flow matching is a deterministic diffusion model with linear interpolation as the forward diffusion process.

From the perspective of the diffusion model, only the solution at the end time-point matters, \ie the generated sample.
This is distinctly different from how general neural \acp{ODE} are used, where the solution is the solving trajectory itself.
Our work is inspired by the recent work of dynamic appearance \acp{ODE} \cite{liuNeuralDifferentialAppearance2024}, the solution trajectory of which is the generated video.
It indicates that the \ac{ODE} flow continuously models the video signal.
However, since they simulate the \acp{ODE} to optimize the textural statistics loss over the trajectory, the training efficiency is limited and the target video must be textures that have spatially stationary visual statistics. Instead, we train our video \acp{ODE} based on the reformulated flow matching, which supports general videos and computationally efficient training.

\mysection{Discussion}{Discussion}
\paragraph{Solve backward}
The \ac{ODE} flow field is invertible.
By instantiating the video trajectory as an \ac{ODE} flow, it enables us to solve the trajectory either forward or backward in time, \ie we can discover not only the future but also the past given one frame.
Note that the diffusion model does not naturally possess this bidirectional capability: For instance, \method{ConDiff} requires an additional flag to condition the solving direction explicitly and swap training data correspondingly to support an analogous functionality \cite{nikankinSinFusionTrainingDiffusion2023}.

Specifically, we can solve backward in time by solving the reverted flow as:
\begin{equation}
\label{eq:joint_solve_backward}
\infSample^{\frameSub-1} = \solve_{1\to 0}(\infSample^{\frameSub} + \noiseLevel\noiseSample, \odej),
\end{equation}
where $(\timeCoord_0, \noiseCoord_0)=(0, 0)$ and $(\timeCoord_1, \noiseCoord_1)=(1, \noiseLevel)$.
A video by solving our \method{Bi-flow} backward is shown in \refFig{SolveBackward}, with the same quality as forward-solved videos except that it is generated in the opposite direction.

\paragraph{Be continuous in time}
With a video \ac{ODE} flow, we can actually model a continuous video trajectory through integration.
Currently, the intermediate states in our solution trajectory are simply linear mixes of the two consecutive frames as it is trained to do so, but to our best knowledge, we are the first method to achieve general video generation in decent quality as solving the \ac{ODE} flow in the same manner of solving initial value problems.

Rather than linear interpolation, any non-linear interpolation in flow matching is possible and actively explored in the community now \cite{liuFlowStraightFast2022,albergoBuildingNormalizingFlows2023}.
It is a promising future direction to obtain infinite temporal resolution in our video flow that we drop in a perceptually interpolating curve during training such that a state at any time point in the solved trajectory is a valid frame.

\paragraph{Limitations}
One of the main limitations of our method is the context size.
Admittedly, our method cannot achieve state-of-the-art quality in video generation, which we mainly attribute to our Markovian sampling where each subsequent frame is generated solely based on the preceding frame.
The context size of only one results in artifacts and incoherent generation, \eg clouds moving in different directions after a few frames and the background gradually morphing to another scene.
The small context also complicates flow training with more crossing and ambiguous data involved.
A promising direction to resolve this is to straightforwardly apply autoregressive sampling to our method.
We leave this for future work.

Another limitation is that we need to manually select the noise level applied in the inference.
This relates to the absence of a reliable estimate of how much the frame has corrupted.
Although we can quickly trial and see to tune this parameter after training, developing an adaptive strategy to add an adequate but minimal amount of noise could be highly beneficial.
For example, we can apply different levels of noise per frame.

\mysection{Conclusion}{Conclusion}
We propose a new paradigm of video generation, \ie to learn an \ac{ODE} flow field that drives from one frame to another.
With our proposed reformulated flow matching and bi-linearity training, we train a joint flow field of temporal evolution and denoising, and in inference, we generate the video in a streaming manner while correcting the frame simultaneously.
Through extensive evaluations, our method presents better efficiency and stability than the widely used conditional diffusion with comparable quality, lending itself to real-time applications such as interactive video creation.

In future work, we would like to extend our bi-flow design beyond the temporal domain to capture other correlations among image distributions, such as multi-view images that introduce a new dimension of camera angles and multi-illumination images that represent a new dimension of lighting conditions.

\small{
\paragraph*{Acknowledgments}
This project was supported by Meta Reality Labs, Grant Nr. 583589.
Chen is further supported by an award from The Rabin Ezra Scholarship Trust.
We also thank the constructive comments from the anonymous reviewers and early discussions with Xingchang Huang. 
}

\setlength{\bibsep}{0.0pt}
{
    \small
    \bibliographystyle{ieeenat_fullname}
    \bibliography{main}
}

\end{document}


\input{project-commands}

\makeatletter
\newcommand{\subalign}[1]{%
  \vcenter{%
    \Let@ \restore@math@cr \default@tag
    \baselineskip\fontdimen10 \scriptfont\tw@
    \advance\baselineskip\fontdimen12 \scriptfont\tw@
    \lineskip\thr@@\fontdimen8 \scriptfont\thr@@
    \lineskiplimit\lineskip
    \ialign{\hfil$\m@th\scriptstyle##$&$\m@th\scriptstyle{}##$\hfil\crcr
      #1\crcr
    }%
  }%
}
\makeatother

\title{Generative Video Bi-flow: Supplemental Materials}

\author{Chen Liu\\
University College London\\
{\tt\small chen.liu.21@ucl.ac.uk}
\and
Tobias Ritschel\\
University College London\\
{\tt\small t.ritschel@ucl.ac.uk}
}
\maketitle

\appendix

We provide more details about our implementation in \refSec{Implementation} and additional quantitative results in \refSec{Quantitative}.
We recommend visiting our supplemental HTML material (\texttt{index.html}) for more generated videos and qualitative evaluations.

\mysection{Implementation Details}{Implementation}
Our implementation uses PyTorch \cite{anselPyTorch2Faster2024}.
We model the ODE flow field using the UNet implementation from Huggingface Diffusers \cite{vonplatenDiffusersStateartDiffusion2024}.
For \method{Flow} and \method{ConDiff}, the UNet network has seven levels of convolutional blocks, with channel sizes of 192, 192, 384, 384, 384, 768, and 768, respectively, and spatial attention layers involved in the penultimate level.
For \method{Bi-flow}, the joint field consists of two separate smaller UNet networks of the same design, except channel sizes being 128, 128, 256, 256, 256, 512, and 512 now.
While we opt not to use a unified model or share parameters among the sub-networks---thereby avoiding the tedious process of tuning loss weights for the two joint losses---we still consider them as a single model to use, rather than as disjoint components.
To ensure fair comparisons, we guarantee that all models have nearly equivalent efficiency and memory footprints, as demonstrated in \refTab{benchmark}.

\begin{table}[!htp]\centering
\caption{We benchmark the models for different methods in terms of the number of parameters (memory) and GFlops (efficiency). GFlops is measured in a $128^2$ input. }\label{tab:benchmark}
\begin{tabular}{lrrr}\toprule
Models &\#Params (M) &GFlops \\\midrule
\method{Flow} &277.867 &134.48 \\
\method{Bi-flow} &247.693 &119.61 \\
\method{ConDiff} &277.872 &134.56 \\
\bottomrule
\end{tabular}
\end{table}

\myfigure{Characteristic}{Characteristic ODE. The orange curve is the characteristic curve parameterized by $\jointCoord$ in the $\timeCoord\noiseCoord$-plane. The solutions of the PDE form a manifold, over which the green curve is the corresponding ODE solving trajectory of this characteristic curve. The tangent of the green curve is the black arrow in the center, with the tangent of the orange curve as the first two dimensions and the directional derivative of $\sample$ as the third dimension.}

The training details are the same across all datasets.
We train all methods using the default AdamW optimizer with a learning rate of $1\times 10^{-4}$.
The batch size is 128, and the number of training iterations is 200K.
We perform gradient accumulation and low-precision training of ``BF16'' using Huggingface Accelerate \cite{accelerate} to fit the large batch size into the limited GPU memory.
Our training operates directly in RGB space.
With the above setting, the average training time is approximately 80 hours in one Nvidia RTX 4090 GPU with PyTorch compilation.
We use Torchode \cite{lienen2022torchode} to solve the learned flow field.

To further explain our joint sampling, in \refFig{Characteristic} we provide an illustration of the characteristic ODE curve (Eq. 8 in the main paper).

\mysection{More Quantitative Results}{Quantitative}

\begin{table*}[!htp]\centering
\caption{The \ac{FID} and I-SIM evaluations for all methods and datasets. The methods and variants are solved using the same configuration of the adaptive solver in the main paper.}\label{tab:FID-ISIM}
\resizebox{\textwidth}{!}{
    \begin{tabular}{lrrrrrrrrrrrrr}\toprule
        &\multicolumn{2}{c}{\dataset{SKY}} &\multicolumn{2}{c}{\dataset{BIKING}} &\multicolumn{2}{c}{\dataset{RIDING}} &\multicolumn{2}{c}{\dataset{CARLA}} &\multicolumn{2}{c}{\dataset{MAZES}} &\multicolumn{2}{c}{\dataset{MINERL}} \\
        \cmidrule(lr){2-3}\cmidrule(lr){4-5}\cmidrule(lr){6-7}\cmidrule(lr){8-9}\cmidrule(lr){10-11}\cmidrule(lr){12-13}
        &\lessIsBetter{FID} &\GreaterIsBetter{I-SIM} &\lessIsBetter{FID} &\GreaterIsBetter{I-SIM} &\lessIsBetter{FID} &\GreaterIsBetter{I-SIM} &\lessIsBetter{FID} &\GreaterIsBetter{I-SIM} &\lessIsBetter{FID} &\GreaterIsBetter{I-SIM} &\lessIsBetter{FID} &\GreaterIsBetter{I-SIM} \\\midrule
        \method{Flow} &246.3 &\winner{0.999} &337.6 &\winner{0.997} &282.7 &\winner{0.997} &287.3 &\winner{0.997} &283.9 &\winner{0.996} &228.0 &\winner{0.998} \\
        \method{Bi-flow} ($\noiseLevel=0.0$) &252.9 &0.998 &307.9 &0.993 &322.8 &0.993 &285.7 &0.993 &214.0 &0.995 &214.8 &0.998 \\
        \method{Bi-flow} ($\noiseLevel=0.1$) &\winner{70.5} &0.961 &211.1 &0.956 &115.5 &0.939 &46.1 &0.943 &97.7 &0.928 &89.3 &0.941 \\
        \method{Bi-flow} ($\noiseLevel=0.2$) &76.7 &0.951 &141.7 &0.918 &83.6 &0.916 &29.1 &0.931 &32.6 &0.883 &53.9 &0.900 \\
        \method{Bi-flow} ($\noiseLevel=0.3$) &80.6 &0.948 &\winner{121.0} &0.897 &\winner{70.3} &0.903 &\winner{26.4} &0.928 &\winner{29.8} &0.879 &\winner{44.4} &0.882 \\
        \method{ConDiff} &191.7 &0.981 &249.5 &0.888 &258.7 &0.929 &124.9 &0.929 &30.2 &0.881 &140.5 &0.859 \\
        \bottomrule
    \end{tabular}
}
\end{table*}

Besides the main discussion of video quality versus speed, we report \ac{FID} for realism and I-SIM for consistency to further investigate frame quality.
\ac{FID} is computed between all 65,536 generated frames (128 videos of 512 frames) and all frames in the test set.
I-SIM is the average cosine similarity of Inception features between consecutive frames of generated videos.
The closer the I-SIM is to one, the smaller the changes between frames.
The numbers are shown in \refTab{FID-ISIM}.

As expected, \method{Flow} wins in terms of consistency, closely followed by \method{Bi-flow} without noise added ($\noiseLevel=0.0$).
This is because the video trajectory modeled by the ODE flow is inherently continuous and coherent, as also evidenced in recent work \cite{liuNeuralDifferentialAppearance2024}.
For \method{Bi-flow} there is an obvious trend that \ac{FID} and I-SIM metrics both decrease with increasing noise levels. 
The decline is particularly pronounced from a noiseless condition ($\noiseLevel=0.0$) to one with noise ($\noiseLevel=0.1$).
This confirms our observation in the main paper that this is a trade-off spectrum where we trade frame quality with frame consistency.
The lowest \ac{FID} is always achieved by \method{Bi-flow}.

\begin{table}[!htp]\centering
\caption{More baseline comparisons in \dataset{SKY} and \dataset{CARLA} datasets.}\label{tab:RIVER_SVP}
    \begin{tabular}{ccc}\toprule
    \method{Bi-flow} &\method{RIVER} &\method{SVP} \\\midrule
    \winner{436} &794 &909 \\
    \bottomrule
    \end{tabular}
\end{table}

In \refTab{RIVER_SVP}, we compare to RIVER \cite{davtyan2023efficient} and SVP \cite{ostrek2024stable} in FVD.
We train them in pixel space for comparison.
RIVER is implemented with the official codebase and the finetuned warm-up hyperparameter.
For SVP, we only adopt the modified inference as their other contributions focus on portrait video generation. 
It shows that our bi-flow outperforms them.
With as few solving steps as bi-flow, both RIVER and SVP accumulate errors rapidly, especially when generating frames beyond the training horizon, while RIVER performs better than SVP due to the sparse conditioning.
It requires approximately $2.53\times$ and $2.69\times$ steps for RIVER and SVP to achieve similar FVD as bi-flow.

We further verify that the sparse conditioning of past frames, one of RIVER's contributions, can also be applied to our bi-flow and improve FVD from 436 to 387 (11\%) by conditioning on one more random previous frame.
We believe that it is orthogonal to our method and shows a promising future direction.

\myfigure{scalability}{Scalability test.}

We show the scalability of our bi-flow in model size and training data in \refFig{scalability}.
For model size, we report the FVD achieved by four variants of the same backbone model, with relative sizes approximately $0.5\times$, $1.0\times$ (original), $2.0\times$, and $4.0\times$.
To demonstrate data scaling, we report FVD on the same test set using random subsets of the training data at 25\%, 50\%, 75\%, and 100\% (original).
The results confirm that our bi-flow can benefit from increasing model capability and data volume.
We have not yet observed saturation within the limit of our available compute resources.

\begin{table}[!htp]
\centering
\caption{UCF101 FVD results.}\label{tab:UCF101}
\begin{tabular}{ccccc}\toprule
\method{Bi-flow} &\method{Flow} &\method{ConDiff} &\method{RIVER} &\method{SVP} \\\midrule
\winner{4512} &5666 &5370 &4794 &5218 \\
\bottomrule
\end{tabular}
\end{table}

To make our evaluation more systematic, we add FVD results of UCF101 \cite{soomroUCF101Dataset1012012}, a multi-class video datasets, in \refTab{UCF101}.
The FVD results reflect the difficulty that UCF101 is significantly more challenging, particularly for unconditional generation.
Video bi-flow continues to outperform the baselines in this situation.

\setlength{\bibsep}{0.0pt}
{
    \small
    \bibliographystyle{ieeenat_fullname}
    \bibliography{main}
}

%% file: tr-commands.tex
\usepackage{soul}
\usepackage{amsmath}
\usepackage{amssymb}
\usepackage{wrapfig}
\usepackage{booktabs}
\usepackage{multirow}
\usepackage{xspace}
\usepackage{xcolor}
\usepackage{tabularx}

\def\figurePath{images/}

\NewDocumentCommand{\rot}{O{45}
O{1em}m}{\makebox[#2][l]{\rotatebox{#1}{#3}}}%

\def\myfigure#1#2{%
    \begin{figure}[htb]%
    \centering\includegraphics*[width = \linewidth]{\figurePath#1}%
    \vspace{-.3cm}%
    \caption{#2}%
    \vspace{-.3cm}%
    \label{fig:#1}%
    \end{figure}%
}

\def\mycfigure#1#2{%
    \begin{figure*}[htb]%
    \centering\includegraphics*[width = \linewidth]{\figurePath#1}%
    \vspace{-.3cm}%
    \caption{#2}%
    \vspace{-.4cm}%
    \label{fig:#1}%
    \end{figure*}%
}

\newcommand{\argmin}[1]{\underset{#1}{\operatorname{arg\,min}\ }}

\newcommand{\refSec}[1]{Sec.~\ref{sec:#1}}
\newcommand{\refFig}[1]{Fig.~\ref{fig:#1}}
\newcommand{\refEq}[1]{Eq.~\ref{eq:#1}}
\newcommand{\refTab}[1]{Tab.~\ref{tab:#1}}
\newcommand{\refAlg}[1]{Alg.~\ref{alg:#1}}

\soulregister\ref7
\soulregister\cite7
\soulregister\refFig7
\soulregister\refAlg7
\soulregister\cite7
\soulregister\ref7
\soulregister\pageref7
\soulregister\shortcite7
\soulregister\eg0
\soulregister\ie0
\soulregister\etal0

\DeclareGraphicsExtensions{.png,.jpg,.pdf,.ai,.psd}
\newcommand{\mysection}[2]{\vspace{-.1cm}\section{#1}\label{sec:#2}\vspace{-.1cm}}
\newcommand{\mysubsection}[2]{\subsection{#1}\label{sec:#2}\vspace{-.1cm}}

\definecollection{mymaths}
\newcommand{\mymath}[2]{
    \newcommand{#1}{\TextOrMath{$#2$\xspace}{#2}}
    \begin{collect}{mymaths}{}{}{}{}
    #1
    \end{collect}
}

\definecolor{colorA}{HTML}{4285f4}
\definecolor{colorB}{HTML}{ea4335}
\definecolor{colorC}{HTML}{fbbc04}
\definecolor{colorD}{HTML}{34a853}
\definecolor{colorE}{HTML}{ff6d01}
\definecolor{colorF}{HTML}{46bdc6}
\definecolor{colorG}{HTML}{000000}
\definecolor{colorH}{HTML}{777777}
\definecolor{colorI}{HTML}{bdd6ff}
\definecolor{colorJ}{HTML}{6a9e6f}

\usepackage{pifont}

\newcounter{fact}

%% file: project-commands.tex
\begin{acronym}
\acro{FVD}{Fr\'echet video distance}
\acro{FID}{Fr\'echet inception distance}
\acro{IADB}{iterative $\alpha$-(de)blending}
\acro{ODE}{ordinary differential equation}
\acro{PDE}{partial differential equation}
\end{acronym}

\mymath{\parameters}{\theta}
\mymath{\ode}{f_\parameters}
\mymath{\odev}{f_\parameters^{\mathrm v}}
\mymath{\oden}{f_\parameters^{\mathrm n}}
\mymath{\odej}{f_\parameters^{\mathrm j}}
\mymath{\distOne}{\mathcal X_0}
\mymath{\distTwo}{\mathcal X_1}
\mymath{\timeCoord}{t}
\mymath{\noiseCoord}{\alpha}
\mymath{\jointCoord}{k}
\mymath{\sample}{\mathbf x}
\mymath{\noiseSample}{\mathbf n}
\mymath{\sourceSample}{\sample_0}
\mymath{\targetSample}{\sample_1}
\mymath{\normalDistribution}{\mathcal N}
\mymath{\uniformDistribution}{\mathcal U}
\mymath{\numberOfFrames}{n}
\mymath{\frameSub}{i}
\mymath{\infSample}{\hat{\sample}}
\mymath{\sourceInfSample}{\infSample_0}
\mymath{\targetInfSample}{\infSample_1}
\mymath{\solveCoord}{s}
\mymath{\solve}{\mathcal{S}}
\mymath{\pair}{\mathcal P}
\mymath{\noiseLevel}{\epsilon}

\newcommand{\task}[1]{\textsc{#1}}
\newcommand{\method}[1]{{\texttt{#1}}}
\newcommand{\mypara}[1]{\noindent\textbf{#1:}\quad}
\newcommand{\dataset}[1]{\texttt{#1}}
\renewcommand{\method}[1]{\textsc{#1}}
\newcommand{\pattern}[1]{\textsc{#1}}

\renewcommand{\eg}{\textit{e.g.}, }
\renewcommand{\ie}{\textit{i.e.}, }
\renewcommand{\wrt}{w.r.t.\ }

\newcommand{\winner}[1]{\textbf{#1}}

\newcommand{\lessIsBetter}[1]{\multicolumn1c{#1 $\downarrow$}}
\newcommand{\GreaterIsBetter}[1]{\multicolumn1c{#1 $\uparrow$}}